\documentclass{article}
\usepackage{spconf,amsmath,epsfig}
\usepackage{amssymb}
\usepackage{spconf,amsmath,epsfig,subcaption}
\usepackage{multirow}
\usepackage{cite}
\usepackage{color}
\let\OLDthebibliography\thebibliography
\renewcommand\thebibliography[1]{
  \OLDthebibliography{#1}
  \setlength{\parskip}{0pt}
  \setlength{\itemsep}{0pt plus 0.3ex}
}

\begin{document}\sloppy

\def\x{{\mathbf x}}
\def\L{{\cal L}}

\title{Two-Stream Joint Matching based on Contrastive Learning for Few-Shot Action Recognition}
%
\name{Long Deng, Ziqiang Li, Bingxin Zhou, Zhongming Chen, Ao Li and Yongxin Ge}
\address{Chongqing University, Chongqing, China}

\maketitle

\begin{abstract}
Although few-shot action recognition based on metric learning paradigm has achieved significant success, it fails to address the following issues: \textbf{(1)} inadequate action relation modeling and underutilization of multi-modal information; \textbf{(2)} challenges in handling video matching problems with different lengths and speeds, and video matching problems with misalignment of video sub-actions. To address these issues, we propose a \textbf{T}wo-\textbf{S}tream \textbf{J}oint \textbf{M}atching method based on contrastive learning (\textbf{TSJM}), which consists of two modules: \textbf{M}ulti-modal \textbf{C}ontrastive \textbf{L}earning Module (\textbf{MCL}) and \textbf{J}oint \textbf{M}atching \textbf{M}odule (\textbf{JMM}). The objective of the MCL is to extensively investigate the inter-modal mutual information relationships, thereby thoroughly extracting modal information to enhance the modeling of action relationships. The JMM aims to simultaneously address the aforementioned video matching problems. The effectiveness of the proposed method is evaluated on two widely used few shot action recognition datasets, namely, SSv2 and Kinetics. Comprehensive ablation experiments are also conducted to substantiate the efficacy of our proposed approach.
\end{abstract}
\section{Introduction}
\label{sec:intro}

Action recognition has achieved significant success in various domains due to the advancement of deep learning~\cite{feichtenhofer2019slowfast},~\cite{lin2019tsm},~\cite{arnab2021vivit}. However, these methods heavily rely on large amounts of training and annotated data, requiring substantial human, financial, and time resources to collect such labeled data, which can be particularly challenging in certain domains. Hence, the emergence of few-shot action recognition aims to address this issue by leveraging a small amount of labeled data to train models that can be rapidly applied to new categories.

\begin{figure}[t]
	\centering
	\includegraphics[width=\linewidth]{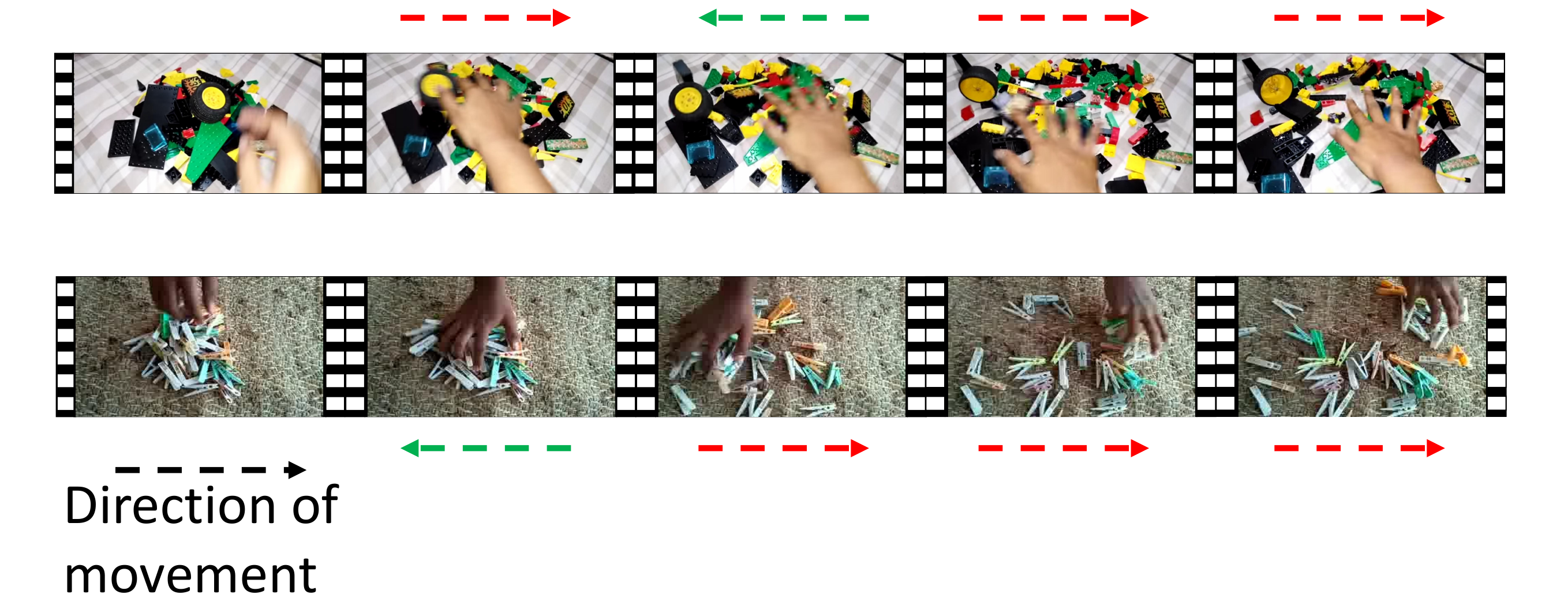}
	\caption{In the equivalent category of “Spreading something onto something", the fundamental behavioral actions remain consistent. However, the agents executing these actions are entirely distinct, represented by toy building blocks and a clamp. Although behavioral actions demonstrate fundamental similarities, an observable misalignment phenomenon arises in the manifestation of sub-actions, accompanied by a distinct reversal in the direction of motion.}
	\label{intro}
\end{figure}

Few-shot action recognition involves more complex structures and additional temporal dimensions compared with few-shot image classification. Majority of the videos on the SSv2~\cite{goyal2017something} dataset are not related to the subjects but strongly correlated with the actions, making action relationship modeling particularly important (Fig.~\ref{intro}). In previous approaches, numerous studies have utilized extra modal information for motion modeling, such as depth information~\cite{fu2020depth}, motion vectors~\cite{luo2022long}, temporal gradients~\cite{xiao2022learning}, frame differences~\cite{wang2023molo} and others. However, the utilization of optical flow information is seldom observed in few-shot action recognition. Optical flow information, which characterizes the direction and speed of object motion by constructing a motion field, holds distinct advantages in modeling motion relationships~\cite{luo2021action},~\cite{wang2017untrimmednets}. Pioneeringly, AMFAR\cite{wanyan2023active} innovatively incorporated optical flow information into the modeling of action relationships through an active learning approach. However, its implementation necessitated supplementary supervised annotation, and it employed a computationally expensive I3D pre-trained model for optical flow learning. In contrast to the aforementioned methodology, our approach leverages contrastive learning to elucidate mutual information across modalities, facilitating the integration of optical flow information into motion modeling without the need for resource-intensive label annotations and parameter-heavy networks.



After effective feature representations are obtained, majority of existing few-shot action recognition methods adopt the paradigm of metric learning to assess the similarity between two videos. These approaches first map video data into a feature space and then utilize various distance metrics to quantify the distance between the support set and the query set, such as dynamic time warping(DTW)~\cite{cao2020few}, Sinkhorn distance~\cite{lu2021few}, and bidirectional mean Hausdorff metric~\cite{wang2022hybrid}. Despite the significant success achieved by these methods in few-shot action recognition, they still fail to address two crucial challenges: \textbf{(1)} the video matching problem with different speeds and lengths, and \textbf{(2)} the issue of video sub-action misalignment. To address the former challenge, OTAM~\cite{cao2020few} utilized the DTW algorithm to dynamically align two videos, effectively mitigating issues associated with variations in speed and length. However, as the complexity of the video increases with the presence of multiple sub-actions, this approach falls short in addressing the latter issue. Specifically, the naive application of the DTW algorithm is susceptible to misaligning two videos sharing the same category, primarily due to disruptions caused by an asymmetric number of sub-actions. Consequently, a paramount objective is to rectify the misalignments arising from the presence of sub-actions in both videos.


We propose a novel two-stream joint matching method based on contrastive learning (\textbf{TSJM}), which comprises two modules: the multi-modal contrastive learning module (\textbf{MCL}) and the joint matching module (\textbf{JMM}), to overcome the above two limitations and enhance the accuracy of few-shot action recognition. Specifically, we first utilize a pre-trained deep convolutional network to extract the appearance features of the objects from RGB video frames and their motion characteristics utilizing optical flow images. Subsequently, grounded on the features extracted from these two modalities, we introduce a novel MCL module. The proposed MCL module is designed to augment the inter-modality interaction, thereby enhancing the acquisition of motion-related patterns within the RGB modality branch. Furthermore, to address the video matching problem, we adopt the DTW algorithm~\cite{muller2007dynamic} to measure the distance between the query sets and the support sets, achieving dynamic alignment of the two video sequences. We also introduce the weighted bipartite graph perfect matching method into few-shot action recognition to address the issue of video sub-action misalignment and utilize the Kuhn-Munkres algorithm~\cite{kuhn1955hungarian} to find the optimal matching. Finally, we combine the scores obtained from the two metric methods to derive the final video similarity, which is utilized for classification. Experimental results demonstrate that our approach yields competitive results on two widely used benchmark datasets.


In summary, the contributions of this work can be summarized as follows:

(1) We introduced a multi-modal contrastive learning module, incorporating the optical flow modality into the few-shot action recognition domain to construct temporal action relationships. Furthermore, we first established the inter-modality mutual information between RGB video frame features and optical flow features through contrastive learning, resulting in a superior representation of videos.

(2) We proposed a joint matching module, where we introduced the weighted bipartite graph perfect matching problem to the few-shot action recognition domain. To address the issue of matching errors caused by video sub-action misalignment, we utilized the KM algorithm to compute the optimal matchings between videos.

(3) Extensive experiments were conducted to demonstrate the effectiveness of our approach, achieving competitive results on two widely used benchmark datasets, SSv2 and kinetics. Moreover, several ablation studies validated the effectiveness of the various modules proposed in this work.

\section{Related Work}



    \noindent\textbf{Motion Informatics Learning}

In the domain of few-shot action recognition, motion information plays a crucial role. Several methods have been proposed for modeling motion information. AMeFu-Net~\cite{fu2020depth} proposed utilizing depth information to assist in modeling object motion. LSTC~\cite{luo2022long} pointed out that motion vectors contain rich temporal relationships, which can demonstrate the motion information of each frame and short-term temporal dependencies. Xiao et al.~\cite{xiao2022learning} suggested that time gradients carry abundant motion signals and proposed using time gradients as an additional modality for motion modeling. AMFAR~\cite{wanyan2023active} used optical flow information as an extra modality to build motion information but utilized a 3D convolutional neural network model, resulting in excessive model parameters that are unsuitable for few-shot action recognition scenarios. In contrast to the aforementioned approaches, we further incorporate contrastive learning to fully explore the relationships between modalities and construct inter-modality mutual information, thereby obtaining more robust feature representations.

    \noindent\textbf{Few-shot Action Recognition}

Few-shot action recognition introduces an additional temporal dimension compared with few-shot image classification, making it more challenging. Prominent studies in this area include OTAM~\cite{cao2020few}, where the DTW algorithm is utilized to measure distances between video frames and find the optimal alignment path. TRX~\cite{perrett2021temporal} replaced single images with binary or ternary tuples to compare subsequences of query and support set videos, addressing the issue of poor matching performance caused by varying speeds and lengths. CMOT~\cite{lu2021few} introduced the optimal transport problem into few-shot action recognition and use Sinkhorn distance to align videos. We introduce the concept of a perfect bipartite matching to few-shot action recognition based on the DTW algorithm, forming a JMM that addresses various matching challenges simultaneously.

\begin{figure*}[t]
	\centering
	\includegraphics[width=\linewidth]{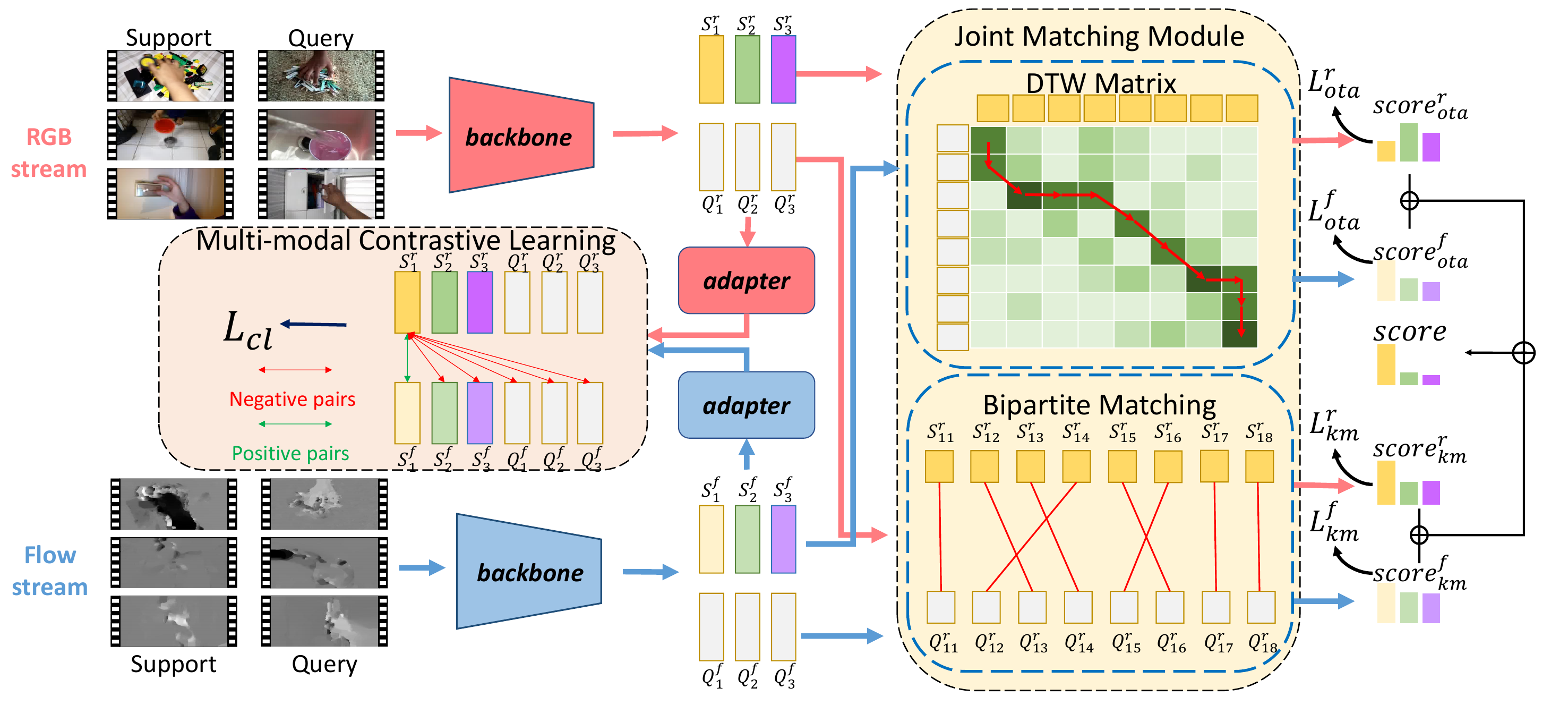}
	\caption{Illustration of the comprehensive framework of our method, taking the 3-way 1-shot task as an example.}
    \label{method}
\end{figure*}

\section{Method}
\subsection{Problem Formulation}
In the domain of few-shot action recognition, the meta-learning paradigm is predominantly utilized to acquire a model with generalization ability, enabling proficient recognition performance on unseen categories. Meta-learning consists of two key phases: meta-training and meta-testing. During the meta-training phase, each training task, denoted as $ \mathcal T$ , is typically sampled from the training dataset $ \mathcal D_{train}$  to form the training samples $ \mathcal T_{train}$ , which include the support set $ \mathcal S$ and the query set $ \mathcal Q$. These training tasks are designed following the N-way K-shot paradigm, where N-way indicates the number of categories in each task, and K-shot represents the number of support set samples per category. In the meta-testing phase, the model, pre-trained through meta-learning, is utilized for inference and generalization on new testing tasks that differ from the meta-training tasks. 

\subsection{Overview}
The overall framework of this work is illustrated in Fig.~\ref{method}. The training samples $ \mathcal T_{train}$ are divided into support and query sets. The 3-way 1-shot is used as an illustrative example. First, a sparse sampling method from TSN~\cite{wang2016temporal} is used to sample video frames and optical flow frames in the support sets and query sets, resulting in T frames. Thereafter, the RGB features $ \mathcal S_{i}^{r}$  and optical flow features $ \mathcal S_{i}^{f}$ of the support sets, and the RGB features $ \mathcal Q_{i}^{r}$ and optical flow features $ \mathcal Q_{i}^{f}$ of the query sets are separately obtained through the feature extractor $\Phi $. Subsequently, an adapter is utilized to bring the multi-modal features closer together, facilitating better contrastive learning. Furthermore the features from both modalities are fed into the JMM for matching, resulting in four scores: $ \mathcal Score_{ota}^{r} $, $ \mathcal Score_{ota}^{f}$, $ \mathcal Score_{km}^{r}$, and $ \mathcal Score_{km}^{f}$. Finally, the final prediction score is obtained through weighted averaging.

\subsection{MCL}

In the same video, the global higher-level feature representations between its two different modalities are interconnected. The mutual information between the two modalities is of crucial importance.

However, the features from different modalities belong to distinct data domains, and their feature disparities are significant, making direct contrastive learning infeasible. Accordingly, we proposes an adapter module. The adapter module is a bottleneck structure comprising two fully connected layers and an activation layer. The first fully connected layer projects the input features to a lower-dimensional space, and the second fully connected layer maps them back to the original dimension. This mechanism enables the projection of two different-domain modality features into a new feature space, facilitating contrastive learning.

After the features of the two modalities are obtained through the feature extractor $\Phi $, they are fed into the adapter module. The RGB modality is expressed as follows:
\begin{equation}
x^{r}_i = \mathcal Adapter(\Phi(X^{r}_i)),X^{r}_i=\left \{ S_{i}^{r},Q_{i}^{r}:(1\le i\le N)\right \} 
\end{equation}
where $\Phi$ represents the feature extractor, and $N$ represents the number of categories.


We utilize a cross-attention mechanism to measure the similarity between global high-level features. Thereafter, we adopt contrastive learning with positive and negative sample pairs. Specifically, we consider the global high-level features from different modalities of the same video as positive sample pairs, while all other combinations form negative sample pairs. Consistent with the approach used in reference CPC~\cite{oord2018representation}, we utilize the InfoNCE loss function.

The specific definition of the contrastive learning loss function $ \mathcal L_{cl}$, is as follows:
\begin{equation}
L_{cl} =-log\frac{ {\textstyle \sum_{i=1}^{k}}exp(sim(x_{i}^{r}, x_{i}^{f})/\tau ) }{{\textstyle \sum_{i\ne j}^{k}}exp(sim(x_{i}^{r}, x_{i}^{f})/\tau ) }
\end{equation}
where $sim(\cdot)$ represents the similarity, and $\tau$ is the temperature hyper-parameter for scaling. The computed similarities are normalized using the “normalize" function and used to construct positive and negative sample pairs.

\subsection{JMM}
We propose a JMM to address the challenges of matching videos with different lengths and velocities and the problem of video sub-action misalignment. This module consists of two sub-modules: ordered temporal matching and bipartite graph matching.

\noindent\textbf{Ordered Temporal Matching}

In this sub-module, we primarily address the issue of matching videos with different lengths and velocities, while preserving the temporal order of video sequences and dynamically aligning them.

We calculate the frame-level distance between the support set feature $ \mathcal S_{i}^{r}$ and the query set feature $ \mathcal Q_{j}^{r}$ of RGB video frames as follows:
\begin{equation}
D_{f} (l,m)=1-\frac{S_{il}^{r} \cdot Q_{jm}^{r} }{\left \| S_{il}^{r} \right \| \cdot\left \| Q_{jm}^{r} \right \|}
\end{equation}
where $D_{f} (l,m)$ represents the frame-level distance between the $l$-th frame of $ \mathcal S_{i}^{r}$ and the $m$-th frame of $ \mathcal Q_{j}^{r}$.

Afterward, we utilize the DTW to compute the shortest distance between the frames of the support sets and the query sets, summing up the optimal match similarity for each frame to obtain the distance between the two videos, $D_{v} (S_{i}^{r},Q_{j}^{r})$:
\begin{equation}
D_{v} (S_{i}^{r},Q_{j}^{r}) =\sum_{j=1}^{T} f(D_{f})
\end{equation}
where $f(\cdot)$ represents the DTW algorithm.

Finally, we optimize the model using the cross-entropy loss as follows:
\begin{equation}
L_{ota}^{r} =-log\frac{exp(-D_{v} (S_{j}^{r},Q_{j}^{r}))}{ {\textstyle \sum_{i=1}^{N}} exp(-D_{v} (S_{i}^{r},Q_{j}^{r}))}
\end{equation}
where $S_{j}^{r}$ and $Q_{j}^{r}$ represent videos from the same category in the support sets and query sets, respectively.


\noindent\textbf{Bipartite Graph Matching}

In this section, we propose a bipartite graph matching module, mapping the concept of bipartite graphs into the context of few-shot action recognition, and utilize the Kuhn-Munkres algorithm within the bipartite graph domain to obtain the maximum weight perfect matching for optimal video alignment.

In the RGB video frames, we calculate the similarity  $sim_{f}(S_{il}^{r},Q_{jm}^{r})$  between the $l$-th frame of the support set feature $S_{i}^{r}$ and the $m$-th frame of the query set feature $Q_{j}^{r}$ using cosine similarity:
\begin{equation}
sim_{f}(S_{il}^{r},Q_{jm}^{r})=\frac{S_{il}^{r} \cdot (Q_{jm}^{r})^{T}  }{\left \| S_{il}^{r} \right \| \cdot\left \| Q_{jm}^{r} \right \|}
\end{equation}
After the similarity is obtained, we treat the support and query set features as nodes in two separate graphs and the similarity between frames as edge weights connecting the nodes. This approach forms a weighted complete bipartite graph between $S_{il}^{r}$ and $Q_{jm}^{r}$, and the objective is to find the maximum weight perfect matching. We utilize the Kuhn-Munkres algorithm for this purpose.

\begin{table*}[htbp]
	\centering
        \setlength{\tabcolsep}{7mm}{
	\begin{tabular}{c|c|c|c|c|c}	\hline
	\multirow{2}{*}{\emph{Methods}}    &\multirow{2}{*}{\emph{Reference}}  &\multicolumn{2}{c|}{\emph{Kinetics}}		&\multicolumn{2}{c}{\emph{SSv2}}\\	

        \cline{3-6}
				                &  &1-shot 	&5-shot 	&1-shot		&5-shot		\\ \hline
		CMN~\cite{zhu2018compound} &ECCV2018	&60.5  	&78.9  	&-  		&- 	\\	
            OTAM~\cite{cao2020few} &CVPR2020	&73.0  	&85.8  	&42.8  		&52.3 	\\
            TRX~\cite{perrett2021temporal} &CVPR2021	&63.6  	&85.9  	&42.0  		&64.6 	\\
            STRM~\cite{thatipelli2022spatio} &CVPR2022	&-  	&86.7  	&-  		&68.1 	\\
            HyRSM~\cite{wang2022hybrid} &CVPR2022	&73.7  	&86.1  	&54.3  		&69.0 	\\
            SloshNet~\cite{xing2023revisiting} &AAAI2023	&-  	&87.0  	&46.5  		&68.3 	\\
            AMeFu-Net~\cite{fu2020depth} &ACMMM2020	&74.1  	&85.8  	&-  		&- 	\\
            LSTC~\cite{luo2022long} &IJCAI2022	&73.4  	&86.5  	&46.7  		&66.7 	\\
            MTFAN~\cite{wu2022motion} &CVPR2022	&74.6  	&\textbf{87.4}  	&45.7  		&60.4 	\\
            MoLo~\cite{wang2023molo} &CVPR2023	&74.0  	&85.6  	&56.6  		&\textbf{70.6} 	\\
            \cline{1-6}
            \textbf{TSJM(ours)} &	&\textbf{75.0}  	&87.0  	&\textbf{58.5}  		&68.5 	\\
		\hline
	\end{tabular}}
	\caption{Comparison with state-of-the-art methods in few-shot action recognition on the Kinetics and SSv2 datasets is conducted. AMeFu-Net, LSTC, MTFAN and MoLo, which are representative multi-modal approaches, incorporate depth information, motion vectors, and inter-frame differences.}
	\label{sota}
\end{table*}

If we find an appropriate feasible vertex labeling, such that the corresponding induced subgraph has a perfect matching, then we obtain the optimal matching for $G$. We then obtain the perfect matching result $M^{*}$ for the query and support set videos, and the final matching score is obtained by summing up the similarity of the matching result, followed by optimizing the model using the cross-entropy loss.

\begin{equation}
sim_{v} (S_{i}^{r},Q_{j}^{r}) = \sum_{j=1}^{T} sim_{f}(M^{*})
\end{equation}

\begin{equation}
L_{km}^{r} =-log\frac{exp(sim_{v} (S_{j}^{r},Q_{j}^{r}))}{\sum_{i=1}^{N} exp(sim_{v} (S_{i}^{r},Q_{j}^{r}))} 
\end{equation}
where $sim_{v} (S_{i}^{r},Q_{j}^{r})$ represents the similarity between the support set video $S_{i}^{r}$ and the query set video $Q_{j}^{r}$. $S_{j}^{r}$ and $Q_{j}^{r}$ indicate the videos of the same category within the support sets and query sets, respectively.

\section{Experiments}
In this section, we carried out a series of experiments on two widely adopted datasets, namely, Kinetics~\cite{carreira2017quo} and Something-Something v2~\cite{goyal2017something}, to empirically validate the efficacy of our approach. We also conducted a comparative analysis between our proposed method and the current state-of-the-art techniques. Additionally, ablation experiments were performed to ascertain the effectiveness of each individual module introduced in this work.

\subsection{Datasets}
We evaluated the effectiveness of our approach on two commonly used few-shot action recognition datasets: Kinetics and Something-Something V2. In both datasets, we utilized a dense optical flow algorithm, similar to the method described in reference~\cite{lucas1981iterative}, to generate optical flow frame sequences from the original videos. In terms of dataset selection, we chose 100 categories from the entire dataset, each containing 100 videos. Consistent with the segmentation approach utilized in references~\cite{cao2020few} and~\cite{zhu2018compound}, we divided the 64 categories into a training set, utilized 12 categories for validation, and allocated the remaining 24 categories for testing purposes.

\subsection{Implementation Details}
In the N-way K-shot setup, we employ a random sampling approach to select N classes, each containing K instances for the support set, while the query set comprises N instances, with each instance belonging to one of the N classes in the support set. During the preprocessing phase, we evenly segment video clips into 8 segments. Subsequently, we randomly select 1 RGB frame from each segment as visual data and choose a pair of consecutive optical flow frames from the same position for motion data. We subject the chosen video frames and optical flow frames to standard data augmentation techniques, resizing them to a 224$\times$224 region. In the RGB video frame branch and the optical flow branch, we utilize ResNet-50~\cite{he2016deep}, pretrained on ImageNet, as the backbone model for both. During training, we utilize the Adam optimizer~\cite{kingma2014adam} with a learning rate of $10^{-5}$. During the meta-test, we follow prior work ~\cite{cao2020few},~\cite{zhu2018compound} to construct 10,000 episodes and report the mean accuracy.
\begin{table}[htbp]
	\centering
        \resizebox{1.0\linewidth}{!}{
	\begin{tabular}{c|c|c}	
            \hline
		\emph{Method}         & \emph{SSv2} 	& \emph{Kinetics} 		\\ 
            \cline{1-3}
		Baseline   	                &42.8  		&73.0 	\\
            Baseline+MCL  	            &52.9  		&73.7 	\\
            Baseline+MCL+adapter 	    &55.3  		&74.1 	\\
            Baseline+JMM  	            &50.9  		&73.8 	\\
            Baseline+MCL+adapter+JMM  	&\textbf{58.5}  		&\textbf{75.0} 	\\
		\hline
	\end{tabular}}
	\caption{Ablation study on SSv2 and Kinetics under 5-way 1-shot.}
	\label{ablation}
\end{table}

\subsection{Comparison with State-of-the-Art Methods}
In this section, we explored the comparative analysis between our proposed method and existing approaches within the field. The comparative experiments will be conducted on datasets widely utilized for few-shot action recognition, maintaining consistency through the adoption of 5-way 1-shot and 5-way 5-shot settings. The chosen comparative methods utilize the pre-trained ResNet-50 from ImageNet as their backbone network to ensure fairness. The experimental results in this study exhibit significant improvements over the baseline (Tab.~\ref{sota}). Furthermore, in comparison to multimodal approaches and the MTFAN method, the 5-way 1-shot experiment on the Kinetics dataset shows an improvement of 0.6\%, while the 1-shot and 5-shot experiments on the SSv2 dataset demonstrate increases of 12.8\% and 8.1\%, respectively. In contrast to the MoLo method, improvements of 1\% and 1.4\% are observed on the Kinetics dataset. The 1-shot experiment on the SSv2 dataset shows an increase of 1.9\%. These experimental results demonstrate the effectiveness of the approach proposed in this work.


\subsection{Ablation Study}
\noindent\textbf{Baseline}
The baseline method employed in this work is first introduced. This method~\cite{cao2020few} is derived from the classic metric-learning-based few-shot action recognition method, utilizing an ImageNet pre-trained ResNet-50 model as the backbone. This method utilizes the DTW algorithm to dynamically align video segments, acquiring the optimal alignment path between the query set and prototypes.






\begin{figure}[htbp]
	\centering
  \resizebox{1.0\linewidth}{!}{
	\includegraphics[width=\linewidth]{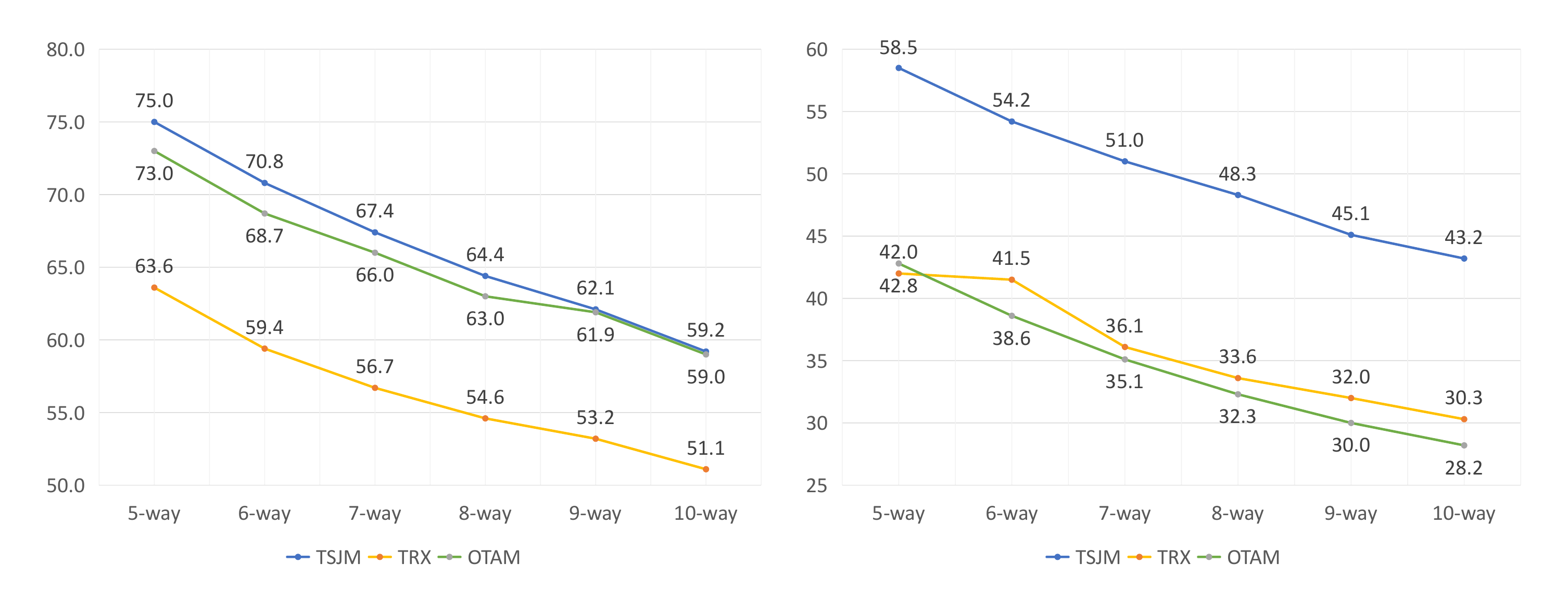}
    }
	\caption{N-way 1-shot on the Kinetics and SSv2.}

	\label{nway}

\end{figure}

\noindent\textbf{Analysis of Network Components}
In this section, an analysis of the efficacy of the individual modules proposed in this work will be conducted. Experimental assessments will be carried out on the SSv2 and Kinetics dataset, employing a 5-way 1-shot experimental setup. The experimental outcomes are presented in Tab.~\ref{ablation}. The experimental data show that the MCL module complements optical flow information with RGB information, resulting in a significant enhancement in accuracy, achieving an improvement of nearly 10 percentage points over the baseline. Meanwhile, the adapter module effectively projects the features of two modalities into a novel feature space, facilitating superior performance in contrastive learning with a notable increase of 2.4\% compared with the absence of the adapter module. In the case of JMM, the introduction of the KM algorithm, in comparison to the baseline, addresses sub-action misalignment issues, resulting in an improvement of 8.1\%. Validation on the Kinetics dataset also attests to the efficacy of each module.

\noindent\textbf{N-way Few-shot Classification}
This section conducts performance evaluations on the SSv2 and Kinetics datasets in the context of N-way 1-shot learning to comprehensively investigate the performance of our proposed methodology under more challenging conditions, where N varies from 5 to 10. An increase in N corresponds to heightened classification complexity and reduced accuracy (Fig.~\ref{nway}). In comparison to prior approaches~\cite{cao2020few},~\cite{perrett2021temporal}, each metric surpasses the corresponding metric of said methods.


\footnotesize
\bibliographystyle{IEEEbib}
\bibliography{icme2023template}

\end{document}